\pdfoutput=1
\documentclass[preprint,12pt]{elsarticle}
\usepackage[a4paper,margin=2.5cm]{geometry}
\usepackage[T1]{fontenc}
\usepackage{lmodern}
\usepackage{amsmath,amssymb}
\usepackage{graphicx}
\usepackage{booktabs}
\usepackage{tabularx}
\usepackage{array}
\usepackage{makecell}
\usepackage{rotating}
\usepackage{xcolor}
\usepackage{microtype}
\usepackage[hidelinks]{hyperref}
\usepackage{xurl}
\usepackage{pdfpages}
\hypersetup{pdftitle={Small-lesion evidence under budgeted whole-slide acquisition: a CAMELYON17 audit with held-out centres},pdfauthor={Iftikhar Ahmed, Adiba Rahman Namira}}
\biboptions{sort&compress}
\graphicspath{{figures/}}

\journal{Computers in Biology and Medicine}

\begin{document}
\begin{frontmatter}
\title{Small-lesion evidence under budgeted whole-slide acquisition: a CAMELYON17 audit with held-out centres\tnoteref{t1}}
\tnotetext[t1]{This is a preprint of a manuscript submitted to \emph{Computers in Biology and Medicine} (Elsevier) on 24 November 2026 and currently under review. It has not been peer reviewed. The supplementary material is appended after the references.}
\author[inst1]{Iftikhar Ahmed\corref{cor1}}
\ead{iftikhar.ahmed@northsouth.edu}
\author[inst1]{Adiba Rahman Namira}
\ead{adiba.namira@northsouth.edu}
\cortext[cor1]{Corresponding author.}
\affiliation[inst1]{organization={North South University}, addressline={Bashundhara}, city={Dhaka 1229}, country={Bangladesh}}
\begin{abstract}
Cost-saving whole-slide analysis encodes only part of each slide at high magnification and is judged by slide-level accuracy. We measured what it loses on lymph node slides with isolated tumour cells (ITC). On 492 CAMELYON17 slides from five centres, in leave-one-centre-out folds over constructed cases, we audited a learned acquisition policy (SASHA, adapted to foundation-model features, three seeds) against scout and random comparators matched per slide in encoder passes, with attention over all tiles as reference. The endpoint was capture of 976 annotated lesions. At 0.375 of full cost, SASHA captured 0.495 (95\% CI 0.336--0.677) of lesions on 29 ITC slides, against 0.884 for attention. Against its own full-slide classifier, SASHA's area under the ROC curve (AUROC) fell from 0.931 to 0.867 overall but from 0.753 to 0.563 for ITC versus negative slides (ITC AUROC), exceeding the pooled loss by 0.126 (post hoc; 95\% CI 0.019--0.237); the direction held in every seed. Under one frozen classifier at the same cost, ITC AUROC fell by 0.08--0.17 for SASHA's tiles and both random comparators, while pooled AUROC fell by 0.02--0.06; a lesion-supervised scout policy's ITC loss was not established. A calibrated fallback rule on SASHA's score released 16/98 small-lesion slides as negative in development. A lesion-supervised scout gatekeeper met three frozen criteria once on 858 held-out same-family slides (1/130 small-lesion slides released; 0.836 of full cost; specificity change $-$0.014), saving mainly through early stops on macrometastases. Evaluations of cost-saving acquisition should report lesion capture by metastasis category at matched total cost.
\end{abstract}
\end{frontmatter}
\section{Introduction}

Weakly supervised multiple instance learning (MIL) on tile features from pathology foundation models now separates metastatic from negative lymph node slides with high accuracy~\cite{ilse2018attention,lu2021clam,chen2024uni}. The price is one encoder pass for every tissue tile at high magnification, about 8,000 tiles per slide on average in the release used here. A line of work therefore acquires only part of each slide, choosing where to look from a low-magnification view or from the evidence gathered so far. SASHA learns a reinforcement learning policy that zooms into 10--20\% of the high-resolution patches~\cite{gogisetty2025sasha}. RLogist and PAMIL learn policies for observation and instance sampling~\cite{zhao2023rlogist,zheng2024pamil}, and PathCTM moves from low to high magnification, stopping once it judges its evidence sufficient~\cite{ge2026pathctm}. These methods report slide-level accuracy or the area under the receiver operating characteristic curve (AUROC) at a fraction of the full encoding cost, and SASHA and PathCTM report parity with analysis at full resolution.

The smallest metastases are the hardest to keep under such a budget. Isolated tumour cells (ITC) are single tumour cells or clusters no larger than 0.2 mm or of fewer than 200 cells. Micrometastases are larger than 0.2 mm but no larger than 2 mm, and macrometastases are larger still~\cite{litjens2018camelyon}. Under the TNM criteria (7th edition) used for the CAMELYON17 labels, a patient whose nodes contain only ITC is staged pN0(i+), and one with micrometastases in one to three axillary nodes is staged pN1mi~\cite{litjens2018camelyon}. The challenge task itself was pN staging at patient level~\cite{bandi2019camelyon17}. On a slide with ITC, the annotated tumour may cover one or two tiles out of thousands. Because larger metastases dominate the positive slides in such cohorts, a pooled slide metric can stay high while the evidence on ITC slides is lost, and reporting at slide level cannot show whether this happens under cost-saving acquisition. A policy trained only on slide labels receives no direct signal to acquire a lesion whose omission rarely changes its training loss.

The closest evaluation study, BagShift, holds the tile features and an MIL predictor fixed while the patch selector changes, and shows that equal patch counts can preserve unequal evidence~\cite{yuan2026bagshift}. Using withheld annotations on CAMELYON16, it showed that a random budget of 128 patches kept annotated tumour in 52.6\% of micrometastatic observations and in 100.0\% of macrometastatic ones. Under that budget, the AUROC for micrometastases against normal slides fell from 0.985 to 0.791, whereas that for macrometastases fell only from 1.000 to 0.994, and micrometastasis sensitivity at a fixed threshold was 63.7\% even for the reference view. BagShift audits stressor selectors on cached full bags at budgets set by patch count, scores retention per slide, has no ITC stratum and no centre hold-out on CAMELYON16, and states that it does not establish whether selector-aware strategies would fail. On CAMELYON17, a segmentation study has reported sensitivity at lesion level for ITC, micrometastases and macrometastases in each held-out centre, without any budget on what is encoded~\cite{almobayed2026pyramid}. To our knowledge, no study has audited a learned acquisition policy at matched total encoding cost, with lesions as units on ITC slides and results reported per centre. Nor has a release rule calibrated to a target for the share of small-lesion slides released as negative, with fallback to full encoding, been checked in each held-out centre.

We measure what cost-saving acquisition loses on ITC and micrometastatic slides, and make four contributions. (i) An audit protocol at lesion level with held-out centres: five leave-one-centre-out folds over constructed cases, annotated lesion components with grid loss kept in the denominator, and capture scored at matched total encoder cost, the low-magnification scout included. (ii) An audit of a learned acquisition policy, SASHA, stratified by ITC and repeated with three training seeds, against scout and random comparators and an attention reference with full information. (iii) A decision analysis at matched cost, stratified by ITC, that extends BagShift's observation on micrometastases to ITC, to learned and scout-guided acquisition, and to held-out centres. Pooled AUROC understated the loss of ITC discrimination at b20 in the primary seed, with paired intervals, for SASHA under its own classifier and, under one frozen predictor, for the slide-supervised and unsupervised policies tested; for a scout policy supervised with lesion labels, the ITC loss was not established. (iv) A calibrated rule that releases slides as negative or falls back to full encoding, evaluated per centre, and a single prespecified confirmation of a scout gatekeeper on 858 held-out slides from the same hospital family. We propose no new selection method, claim no clinical safety and no risk guarantee across hospitals, and do not present the held-out confirmation as external validation.

\section{Related work}

\textbf{Learned and adaptive acquisition.} SASHA observes a slide at low magnification and uses a reinforcement learning policy to choose, one at a time, the regions to acquire at high magnification. A hierarchical attention-based feature distiller (HAFED) aggregates the acquired features, and a targeted state updater (TSU) updates the states of regions that resemble the one just acquired~\cite{gogisetty2025sasha}. With 10--20\% of the high-resolution patches, SASHA matched methods that use every patch on CAMELYON16 and TCGA lung cancer. It also reports four-class CAMELYON+ results (accuracy, AUROC and F1) comparable to such methods, for which minority classes such as ITC were oversampled during training. Its authors further showed that the policy samples patches richer in tumour than random sampling does, followed how quickly it reaches tumour patches on slides with different tumour fractions, examined the slides it misclassified, and noted that at a 10\% budget it tends to misclassify slides with small tumour regions. RLogist learns an observation strategy with deep reinforcement learning~\cite{zhao2023rlogist}, and PAMIL couples instance sampling with a learned policy~\cite{zheng2024pamil}. PathCTM switches dynamically between scales, prunes regions by attention and stops early when confident, reporting 95.95\% fewer patches with AUROC maintained~\cite{ge2026pathctm}. InfoDPP-PAC selects patches by relevance and determinantal diversity and stops with a probably approximately correct (PAC) certificate for the residual information gain; it explicitly does not claim an end-to-end diagnostic improvement~\cite{mittal2026infodpp}. These studies report accuracy or AUROC at slide level together with savings in patches or time, and SASHA also reports how often it hits tumour patches. We found no acquisition study that reports lesion capture by metastasis category, and none that compares a learned policy with cheap comparators at matched total cost, the low-magnification pass included.

\textbf{Evidence retention and model probing.} BagShift treats selection shift as an intervention within each case and measures how a fixed predictor responds to different selectors on PANDA, TCGA lung to CPTAC and CAMELYON16, and it reports AUROC stratified by stage under selection~\cite{yuan2026bagshift}. HIPPO builds counterfactual slides by removing or adding tissue regions, and in lymph node metastasis detection it uncovered model limitations that standard metrics and attention did not reveal~\cite{kaczmarzyk2024hippo}. More generally, a model can fail on clinically important subgroups that its aggregate metrics do not reveal, a problem known as hidden stratification~\cite{oakdenrayner2020hidden}. MHIM-MIL and ACMIL mask the most salient instances during training, either to mine hard instances or to stop attention from concentrating on a few tiles~\cite{tang2023mhimmil,zhang2025acmil}. These are training methods, not audits of acquisition. Outside pathology, visual active search learns policies that find sparse target regions in aerial imagery under a query budget~\cite{sarkar2024vas}, a decision problem close to the one a cost-saving sampler faces on an ITC slide.

\textbf{Calibration and risk control under shift.} Split conformal prediction gives finite-sample guarantees when calibration and test data are exchangeable~\cite{angelopoulos2023gentle}, and conformal risk control extends them to monotone losses such as a false negative rate~\cite{angelopoulos2024risk}. A new hospital breaks exchangeability, and coverage then degrades by an amount that depends on the shift~\cite{barber2023beyond}. Tumour identification across hospitals is a standard example of such shift~\cite{koh2021wilds}. We therefore treat every calibrated threshold here as an empirical rule whose attainment is measured in each held-out centre, not as a guarantee.

\textbf{Validation and reporting.} Metrics Reloaded argues that validation metrics should follow from the properties of the clinical problem~\cite{maierhein2024metrics}. For lymph node staging, one such property is that the smallest category is both the rarest and the hardest. We report the study according to the 2024 Checklist for Artificial Intelligence in Medical Imaging (CLAIM)~\cite{tejani2024claim} (Supplementary Table S8).

\section{Materials and methods}

\subsection{Data and benchmark}

This retrospective study used the CAMELYON+ release~\cite{ling2025camelyonplus,ling2025camelyonplusdata} of the CAMELYON16 and CAMELYON17 sentinel lymph node slides~\cite{litjens2018camelyon,ehteshamibejnordi2017}. The release provides corrected slide labels in four classes (negative, ITC, micrometastasis, macrometastasis), polygon annotations of positive regions, tile coordinates and UNI tile features. The benchmark comprises the 492 slides of the CAMELYON17 training portion whose centre is given in the official stage table~\cite{camelyonaws}. They form 100 cases from five centres, with 20 cases and 98--99 slides per centre; eight cases keep four of their five slides in the release. CAMELYON17 grouped its slides into artificial patients, each from a single centre~\cite{litjens2018camelyon}. We call these groups constructed cases and use them as the statistical unit, never as verified patients. The benchmark holds 318 negative, 36 ITC, 62 micrometastatic and 76 macrometastatic slides. ITC counts as positive in every slide-level analysis, although clinical staging records it as pN0(i+). Three slides keep the corrected CAMELYON+ label (micrometastasis) instead of the original label in the stage table (macrometastasis).

Each of the five folds holds out one centre as its test set, and every case from the other four centres keeps the same development role in all folds. Within each centre, cases were ordered by the smallest lesion class they contain, with ties broken by a seeded hash, and dealt the repeating pattern train, validation, calibration, train, train. This gives 12, 4 and 4 cases per centre, so each fold has 48 training, 16 validation, 16 calibration and 20 test cases (Supplementary Table S1). Roles were assigned from slide labels and identifiers only, and every slide serves as a test slide exactly once. Training cases fit the models, and validation cases select early stopping epochs, checkpoints and thresholds. Calibration cases are never fitted; they serve only the calibrated rules of Section 3.6. The benchmark was frozen before any model was trained, and a dated ledger records every analysis, whether it was prespecified and which outcomes it read (Supplementary Table S2).

The remaining 858 slides of the release, 472 from the CAMELYON17 test portion and 386 from CAMELYON16, have no centre recorded per slide in the official metadata. They never entered development and form the confirmation set of Section 3.7.

\subsection{Reference lesion endpoint}

The lesion endpoint covers the 156 benchmark positives whose annotations passed programmatic checks: 29 ITC, 53 micrometastatic and 74 macrometastatic slides. The other 18 positives (7 ITC, 9 micrometastatic and 2 macrometastatic) have no usable annotation, because 15 have no annotation file and 3 are flagged for geometry review. They stay in every slide-level analysis but in no lesion denominator. Following the CAMELYON+ statement that its annotations outline positive regions~\cite{ling2025camelyonplus}, we read ungrouped and ``metastases'' regions as positive. Each polygon part of the union of a slide's positive regions is one lesion, and overlapping regions merge. This gives 976 lesions: 95 on ITC slides, 372 on micrometastatic slides and 509 on macrometastatic slides.

Under the primary rule, a tile overlaps a lesion when their intersection has positive area. Two sensitivity rules were defined before any capture was computed. Under substantial overlap, the intersection must be at least the smaller of 50\% of the lesion area and 10\% of the tile area. Under merged units, lesions within 512 level-0 pixels (one tile width) of each other merge. Forty-one lesions overlap no candidate tile (3 ITC, 20 micrometastatic, 18 macrometastatic). We call this grid loss. These lesions stay in every denominator as misses, which caps capture at 0.968, 0.946 and 0.965 for the three categories. Lesions on ITC slides overlap a median of 2 candidate tiles (quartiles 1--3).

An acquisition captures a lesion when at least one acquired tile overlaps it, and capture is the proportion of lesions captured, pooled over slides so that each lesion counts equally. Capture is a geometric reference endpoint, not diagnostic sensitivity. The annotations are not exhaustive, no pathologist adjudicated them again, and a captured lesion still has to be recognised by the classifier. The annotation frame was checked in every centre (median covered fraction 1.0) and visually on six slides from four centres whose source images were available locally. One ITC slide (patient\_068\_node\_1) remains in doubt, because its lesions lie mostly outside the tile grid and its source image was not inspected. Section 4.5 reports the effect of excluding it.

\subsection{Features and predictors}

Tiles are 256 $\times$ 256 pixels at pyramid level 1 of the source images (nominally 20$\times$), as tiled by the release. A full encode of a slide costs $N$ encoder passes, one per tissue tile. We used the release's 1,024-dimensional UNI features~\cite{chen2024uni}. Attention ranks and lesion overlaps assume that feature row $i$ belongs to coordinate row $i$. To check this, we extracted UNI features again from the original images with the published checkpoint~\cite{uni_model} for 512 sampled tiles on 12 slides from the five centres, the CAMELYON17 test portion and CAMELYON16. Of these, 511 tiles were assessed and all matched their coordinates; the remaining tile could not be assessed because of an opacity anomaly. This verifies a sample, not every row.

The low-magnification scout encodes each cell of 2,048 $\times$ 2,048 level-0 pixels (4 $\times$ 4 tile positions) once, by applying UNI to the 256-pixel crop at pyramid level 3 (nominally 5$\times$). A scout pass over all $C$ cells of a slide costs $C$, on average 0.098 of $N$.

The common slide predictor is gated attention-based MIL (ABMIL)~\cite{ilse2018attention} trained on full bags, with hidden width 512, attention width 256, dropout 0.25, Adam with learning rate $10^{-4}$ and weight decay $10^{-4}$, at most 50 epochs with patience 10, and seeds 0--4. Its slide score is the mean of the five seeds' probabilities. A logistic regression on mean-pooled features is reported as a weak reference. The hyperparameters of both models were fixed in advance rather than tuned, and validation slides chose only the ABMIL stopping epoch and the thresholds. Thresholds maximise Youden's J~\cite{youden1950index} on the fold's validation slides (rule R0). SASHA's own classifier, HAFED applied to every cell, gives its native decision on the full slide.

\subsection{Acquisition policies at matched total cost}

Cost is counted in encoder passes, one unit per scout cell and one per acquired 20$\times$ tile. A policy that runs the scout and then acquires $R$ tiles costs $T = C + R$, capped at $N$ (the cap never bound on any benchmark slide) and reported as a fraction of $N$. Tissue detection, file reading and the policy's own computation are not counted, so every cost in this paper is an estimate in encoder passes, not a timing.

\textbf{SASHA.} SASHA observes every cell at low magnification, chooses cells to acquire with a reinforcement learning policy, aggregates the acquired features with HAFED, updates similar cells with the TSU and classifies the slide at the end of the episode~\cite{gogisetty2025sasha}. We retrained it per fold on the benchmark's features from the public code~\cite{sasha_code}, using SASHA's CAMELYON16 configuration on binary slide labels and without the oversampling of minority classes used in its four-class CAMELYON+ variant. HAFED was trained for 50 epochs, the TSU for 200 epochs at a 10\% budget and 1,000 at 20\%, and the policy for 15 epochs of proximal policy optimisation with two environments. The cosine threshold was 0.95. The budgets were 10\% and 20\% of cells (b10 and b20). The adapter (Supplementary Table S3) changes three things. First, the low-resolution state is the 1,024-dimensional UNI scout feature instead of a 384-dimensional feature from a small vision transformer (ViT-S). Second, a visited cell acquires the release's tissue tiles whose centres fall inside it (at most 25, zero-padded and masked in HAFED's attention within the cell) instead of 16 fixed sub-patches that include background. Third, calibration cases never enter SASHA's training. Checkpoints were chosen by SASHA's own validation rule. The upstream seed for reinforcement learning was hard-coded; we made it configurable and trained three pairs of stage and reinforcement learning seeds: 4:1130 (the original value) at both budgets, and 5:1135 and 6:1136 at b20. Because a visited cell brings all of its tiles, SASHA acquired 0.143 (b10) and 0.278 (b20) of the tiles on average, more than its nominal cell budget. Its mean total cost was 0.241 and 0.375 of a full encode.

\textbf{Comparators.} Every deployable comparator is matched per slide to SASHA's own cost $T$ at each budget and seed. A standalone policy would have to fix this allowance in advance. Here it is inherited from SASHA, so the comparison isolates where tiles are taken, not how many.

\begin{itemize}
\item \emph{Scout + random cells} runs the scout and opens cells in random order until exactly $R$ tiles are acquired, cutting the last cell in tile order. It separates SASHA's learned ordering from the scout and from the cell geometry (200 Monte Carlo draws per slide, with a standard error of at most 0.0045). A whole-cell variant, which opens the last cell completely, serves as a sensitivity analysis.
\item \emph{Scout-greedy} opens cells in decreasing order of the score of a logistic regression on the 5$\times$ cell feature (L2 penalty, C = 1, balanced class weights) until exactly $R$ tiles are acquired, again cutting the last cell in tile order. The regression was trained per fold on training slides with lesion labels: positive cells overlap an annotated lesion, and negative cells come from negative slides. Scout-greedy thus uses supervision at lesion level that SASHA never receives, which makes it a comparator with stronger supervision rather than a like-for-like one.
\item \emph{Random tiles} acquires $T$ tiles uniformly at random without a scout. Its capture is the exact hypergeometric expectation.
\item \emph{Attention top-R} and \emph{top-T} acquire the $R$ or $T$ tiles with the highest mean ABMIL attention. Ranking by attention requires every tile to be encoded, so their true cost is $N$. They serve as a reference with full information, not as a cost-saving policy.
\end{itemize}

\subsection{Endpoints and prespecified readings}

The primary endpoint, fixed before any SASHA acquisition was scored, is the capture of lesions on the 29 assessable ITC slides at b20 with seed 4, pooled over lesions. It has three paired contrasts: SASHA minus scout + random cells, SASHA minus scout-greedy, and SASHA minus attention top-R. Two readings were fixed in advance. SASHA's policy adds value if the 95\% CI of the first contrast lies above 0. Developing an acquisition method instead of auditing one would be reconsidered if SASHA were close to full information, that is, if the CI of the third contrast included 0, or if the CI of the second contrast lay above 0 with a point estimate of at least 0.15. Secondary endpoints are capture on micrometastatic and macrometastatic slides, capture at b10, capture of lesions that overlap a single candidate tile, and capture per centre.

Slide decisions were scored in two ways. In the native comparison, SASHA's own probability at each budget is compared with HAFED on all cells and with ABMIL on the full bag. The comparison uses AUROC for all positive against negative slides (pooled AUROC), AUROC for ITC against negative slides (ITC AUROC), and calls at the validation Youden threshold. Because the native comparison mixes acquisition with SASHA's own classifier, we also scored the acquired tiles of every policy with one predictor, following BagShift's design with a fixed predictor~\cite{yuan2026bagshift}. This common predictor is each fold's frozen full-bag ABMIL, applied exactly to the acquired tiles, with its attention softmax renormalised over them. The threshold was either refitted per policy on validation slides (T1; for random policies, refitted per draw over 100 draws) or transferred unchanged from the full bag (T2).

The readings of this analysis were fixed before it was run (Supplementary Table S4):

\begin{itemize}
\item whether every achievable policy loses ITC AUROC against the full bag (D1);
\item whether SASHA's tiles lose it under the common predictor (D2);
\item which differences between achievable policies are established (D3), with no ranking stated unless one is;
\item whether the capture gap to attention survives the stricter lesion rules (L1).
\end{itemize}

A table of evidence and calls classifies each assessable ITC slide by whether any of its lesions was captured and whether the slide was called positive, and a transition table pairs each slide's call under a policy with its call on the full bag. A further reading, post hoc, was fixed after an external review but before its interval was computed: the subgroup disparity is established when the 95\% CI of D = (ITC AUROC loss) $-$ (pooled AUROC loss) lies above 0 (R-D; Supplementary Table S4).

\subsection{Calibrated cascade with release or fallback}

A cheap pass gives each slide a score $s$ at cost $T$. The slide stops as positive if $s \geq t_{\mathrm{high}}$ (the validation Youden threshold) and is released as negative if $s < t_{\mathrm{low}}$. Otherwise it falls back to the full encode, at a total cost of $C + N$, and ABMIL on the full bag decides at R0. The threshold $t_{\mathrm{low}}$ is a split conformal threshold~\cite{angelopoulos2023gentle} computed on the fold's 11--16 calibration slides with small lesions (ITC and micrometastatic). With $\alpha$ = 0.10 and $k = \lfloor \alpha (n + 1) \rfloor = 1$, it is the lowest calibration score among these slides, which bounds the probability of releasing an exchangeable small-lesion slide by $k/(n + 1) \leq \alpha$. A new centre breaks exchangeability~\cite{barber2023beyond}, so we measured attainment in each centre rather than assuming it. The primary cheap pass was SASHA at b20. The secondary pass was the scout alone, at cost $C$, scored by the maximum 5$\times$ cell score of the scout scorer trained with lesion labels. A controlled saving was prespecified as a pooled release rate of small-lesion slides of at most $\alpha$ together with an upper 95\% bound on the pooled mean cost below 1.

The scout gatekeeper was defined after these results (Supplementary Table S2). It combines the five fold pipelines. For pipeline $k$, $t_{\mathrm{low},k}$ is defined as above and $t_{\mathrm{high},k}$ is the $j$-th largest score among the calibration negatives ($\beta$ = 0.10, $j$ = 4--5). A slide is released only if all five pipelines release it, and it stops as positive only if all five stop it. Otherwise it falls back to the full encode, where it is called positive if at least three of the five fold ABMIL ensembles reach their R0 thresholds. The comparator, which we call the full pipeline, applies the same majority vote to every slide. A stop or a release costs $C$, and a fallback costs $C + N$. Because $t_{\mathrm{high}}$ is itself a calibration score, the $\beta$ bound for positive stops assumes no ties, so we audited ties. As an exploratory analysis on development folds, a release-only variant turns every positive stop into a fallback, which separates the saving from negative release from the saving from positive stops. Under reading R1, release alone saves encoder passes if the upper bound of the mean cost is below 1.

\subsection{One-time confirmation}

The frozen gatekeeper was run once on the 858 held-out slides (Table 1): 472 slides from the CAMELYON17 test portion (315 negative, 10 ITC, 41 micrometastatic, 106 macrometastatic) and 386 CAMELYON16 slides (238, 8, 71 and 69). Before this run, no confirmation slide had entered training, thresholds, rule choice or any outcome analysis; earlier audits of the data format had opened a few of them without labels or model outputs. Their 5$\times$ scout features were encoded once, for this run. Three primary criteria had to hold, all fixed before any confirmation score was computed:

\begin{itemize}
\item P1: the upper end of the exact two-sided 95\% Clopper--Pearson interval~\cite{clopper1934binomial} for the small-lesion release rate is at most 0.10;
\item P2: the upper 95\% bootstrap bound of the mean per-slide cost fraction is below 0.90;
\item P3: the lower 95\% bound of the paired specificity difference (cascade minus full pipeline) is at least $-$0.05.
\end{itemize}

The statistical units were 486 cases: the constructed CAMELYON17 cases plus one case per CAMELYON16 slide, because the linkage of CAMELYON16 slides to patients is unknown. The CAMELYON17 test slides come from the five benchmark centres, and CAMELYON16 was acquired at two hospitals that also contributed to CAMELYON17~\cite{litjens2018camelyon}. The confirmation therefore uses held-out slides from the same hospital family, with no prior use of their outcomes; it is not external validation.

\subsection{Statistics}

Intervals are 95\% percentile bootstrap intervals~\cite{efron1979bootstrap} over constructed cases resampled within centre, with 2,000 resamples and a fixed seed per analysis (Supplementary Table S6). For the confirmation set, which has no centre per slide, cases were resampled within source (CAMELYON16 or the CAMELYON17 test portion). A resampled case brings all of its slides, and contrasts are paired within resamples. The disparity D was bootstrapped on the same resamples as its two losses, so its interval is paired with theirs. A sensitivity analysis recomputes D only from positive and negative slides of the same held-out centre, which compares only slides scored by the same fold model; pooling these pairs over centres gives more weight to centres with more eligible pairs. Pooled numbers concatenate the out-of-fold predictions of five models, one per fold. Release counts use exact Clopper--Pearson intervals, which treat slides as independent. Capture is an exact expectation for random tiles and a Monte Carlo estimate for random cells, and decisions of random policies average 100 draws with thresholds refitted per draw. The intervals are conditional on the fitted models and the five centres. They do not cover variation between training seeds, which three SASHA seeds address descriptively, or variation between hospitals, for which five centres are too few. We call a contrast established when its 95\% CI excludes 0 and otherwise say that it was not established; an interval that includes 0 is not read as evidence of no difference. All intervals are pointwise and are not adjusted for the number of policies, settings, readers or scales compared. The analysis ledger (Supplementary Table S2) labels each analysis as prespecified on discovery data, post hoc, or confirmatory, and only the confirmation of Section 3.7 is confirmatory. Every benchmark test fold was read many times, and earlier results shaped later choices.

\subsection{Implementation and reproducibility}

The evaluations ran on a CPU in Python 3.12 with NumPy and PyTorch, using shapely for lesion geometry. SASHA's preprocessing ran on a local GPU and its reinforcement learning on cloud T4 GPUs, taking about 1.9 h per fold at b20. Every evaluation records its code commit and the SHA-256 hashes of its inputs and outputs, and it refuses to overwrite an existing output. Code availability is stated in the Data and code availability section.

\section{Results}

Square brackets give 95\% confidence intervals throughout.

\subsection{Cohort, endpoint and missingness}

Table 1 summarises the data and the reference standard. Of the 174 benchmark positives, 156 could be assessed at lesion level. Missing annotations were unevenly spread: 7 of 36 ITC slides had no usable annotation, against 9 of 62 micrometastatic and 2 of 76 macrometastatic slides, and centre 0 held 8 of the 18 positives that could not be assessed. Grid loss was small. Only 3 of the 95 lesions on ITC slides, 20 of 372 micrometastatic lesions and 18 of 509 macrometastatic lesions overlapped no candidate tile. Lesions on ITC slides were the smallest units, overlapping a median of 2 candidate tiles (quartiles 1--3), against 3 (2--7) on micrometastatic and 6 (2--201) on macrometastatic slides. Fig. 1 shows the analysis flow and the boundary between discovery and confirmation.

\begin{table}[!t]
\centering
\small
\caption{Data and reference standard. Benchmark slides come from five CAMELYON17 centres (100 constructed cases); the confirmation set has no slide-level centre (486 statistical units). Per-centre and per-fold counts are in Supplementary Table S1.}
\label{tab:1}
\begin{tabularx}{\textwidth}{@{}>{\raggedright\arraybackslash}X*{5}{r}@{}}
\toprule
 & Negative & ITC & Micro & Macro & Total \\
\midrule
Benchmark slides & 318 & 36 & 62 & 76 & 492 \\
Assessable for the lesion endpoint & --- & 29 & 53 & 74 & 156 \\
Positives without a usable annotation & --- & 7 & 9 & 2 & 18 \\
Reference lesions & --- & 95 & 372 & 509 & 976 \\
Lesions overlapping no candidate tile (grid loss) & --- & 3 & 20 & 18 & 41 \\
Capture ceiling & --- & 0.968 & 0.946 & 0.965 & 0.958 \\
Candidate tiles per lesion, median (quartiles) & --- & 2 (1--3) & 3 (2--7) & 6 (2--201) & --- \\
Confirmation set, CAMELYON17 test portion & 315 & 10 & 41 & 106 & 472 \\
Confirmation set, CAMELYON16 & 238 & 8 & 71 & 69 & 386 \\
\bottomrule
\end{tabularx}
\end{table}

\begin{figure}[!t]
\centering
\includegraphics[width=\textwidth]{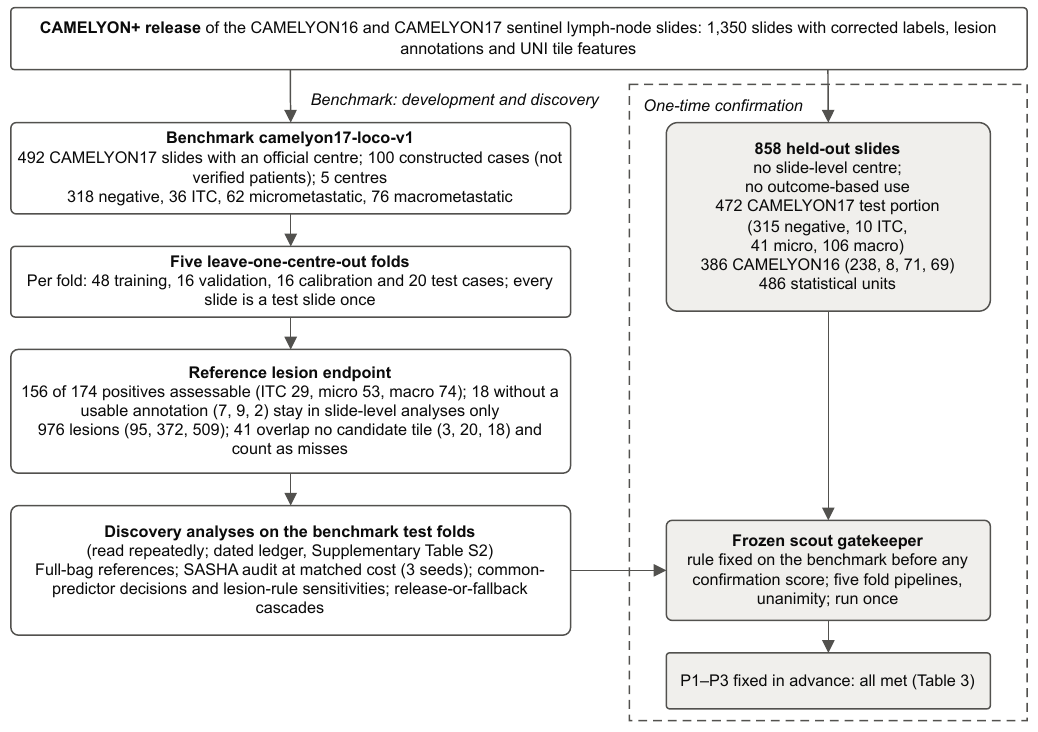}
\caption{Cohort and analysis flow: benchmark roles per fold, exclusions from the lesion endpoint by category, the confirmation set, and the boundary between discovery analyses (benchmark test folds) and the single confirmatory run. Constructed cases are not verified patients. The diagram was drafted with the Claude Opus 5.5 model and rendered with draw.io; the authors checked its content.}
\label{fig:1}
\end{figure}

\subsection{Full-information references}

On the full bag, ABMIL reached a pooled AUROC of 0.942 [0.907, 0.971] over the held-out centres, and the logistic regression on mean-pooled features reached 0.674 [0.617, 0.731]. At R0, ABMIL called 23 of 36 ITC slides, 60 of 62 micrometastatic slides and all 76 macrometastatic slides, at a pooled specificity of 0.720. Its specificity ranged from 0.293 to 0.983 across held-out centres, so validation thresholds transferred poorly between centres. Its ITC AUROC was 0.779 [0.673, 0.881]. Ranking tiles by its attention kept most of the evidence on small lesions. The top 5\% of tiles captured 0.768 [0.644, 0.882] of the lesions on ITC slides, against 0.126 [0.104, 0.160] for a random 5\%, and kept pooled discrimination (AUROC 0.943 [0.907, 0.972]; difference 0.001 [$-$0.004, 0.006]). Its calls moved with the validation threshold: it called 25 rather than 23 ITC slides, and its specificity rose from 0.720 to 0.906. A random 5\%, by contrast, reduced ITC AUROC to 0.508 [0.439, 0.584]. Full information thus keeps most of the ITC evidence and the discrimination of the full bag, and every loss reported below is read against this reachable reference.

\subsection{Lesion capture at matched total cost}

At b20 (mean cost 0.375 of a full encode, of which the scout accounts for 0.098), SASHA captured 0.495 [0.336, 0.677] of the lesions on ITC slides. Scout + random cells captured 0.342 [0.303, 0.387], scout-greedy 0.484 [0.407, 0.586], random tiles 0.599 [0.544, 0.657] and attention top-R 0.884 [0.770, 0.969] (Table 2; Fig. 2a). SASHA's advantage over random cells was not established at b20 (0.153 [$-$0.013, 0.324]) but was established at b10 (0.224 [0.114, 0.321]). Its difference from scout-greedy was 0.011 [$-$0.164, 0.186], and its gap to full information was $-$0.389 [$-$0.575, $-$0.202]. By the prespecified reading, SASHA was not close to full information, so the study remained an audit. On micrometastatic slides, SASHA captured 0.427 [0.277, 0.639] at b20, against 0.742 for scout-greedy and 0.685 for random tiles, and on macrometastatic slides it captured 0.640 [0.539, 0.761]. Lesions that overlap a single candidate tile remained the main gap. SASHA captured 0.26 of them at b10 and 0.34 at b20, against 0.51--0.58 for scout-greedy and 0.68--0.73 for attention top-R across the two budgets. Capture in each centre is given in Supplementary Table S5.

\begin{sidewaystable}
\centering
\scriptsize
\setlength{\tabcolsep}{3.5pt}
\caption{Comparison at matched cost: b20, SASHA seed 4 (mean cost 0.375 of a full encode). Lesion capture comes from the acquisition audit. The decision columns use the common frozen full-bag ABMIL, with validation thresholds refitted for each policy (T1). ITC AUROC compares ITC with negative slides. Capture of random tiles is an exact expectation, and capture of random cells a Monte Carlo estimate over 200 draws; decisions of random policies are means over 100 draws, so their ITC calls are given as rates. The full bag captures every lesion that overlaps a candidate tile, so its capture equals the ceiling.}
\label{tab:2}
\begin{tabular}{@{}>{\raggedright\arraybackslash}p{2.5cm}>{\raggedright\arraybackslash}p{1.95cm}*{9}{c}@{}}
\toprule
Policy & \makecell[l]{Supervision of\\the acquisition\\signal} & \makecell{Needs a\\full\\encode} & \makecell{Mean\\cost / $N$} & \makecell{Capture,\\ITC\\slides} & \makecell{Capture,\\micro\\slides} & \makecell{Capture,\\macro\\slides} & \makecell{Pooled\\AUROC} & \makecell{ITC\\AUROC} & \makecell{ITC\\slides\\called} & \makecell{Specificity} \\
\midrule
Full bag & slide labels & yes & 1.000 & 0.968 & 0.946 & 0.965 & \makecell{0.942\\{[}0.909, 0.969{]}} & \makecell{0.779\\{[}0.669, 0.877{]}} & 23/36 & 0.720 \\[2pt]
Attention top-R (reference) & slide labels & yes & 1.000 & \makecell{0.884\\{[}0.770, 0.969{]}} & \makecell{0.909\\{[}0.799, 0.980{]}} & \makecell{0.853\\{[}0.788, 0.929{]}} & \makecell{0.942\\{[}0.909, 0.970{]}} & \makecell{0.778\\{[}0.669, 0.878{]}} & 24/36 & 0.673 \\[2pt]
SASHA & slide labels & no & 0.375 & \makecell{0.495\\{[}0.336, 0.677{]}} & \makecell{0.427\\{[}0.277, 0.639{]}} & \makecell{0.640\\{[}0.539, 0.761{]}} & \makecell{0.879\\{[}0.834, 0.917{]}} & \makecell{0.608\\{[}0.503, 0.718{]}} & 13/36 & 0.720 \\[2pt]
Scout + random cells (last cell cut) & none & no & 0.375 & \makecell{0.342\\{[}0.303, 0.387{]}} & \makecell{0.415\\{[}0.355, 0.511{]}} & \makecell{0.578\\{[}0.511, 0.681{]}} & \makecell{0.884\\{[}0.844, 0.917{]}} & \makecell{0.615\\{[}0.534, 0.700{]}} & 0.429 & 0.746 \\[2pt]
Scout-greedy & lesion labels & no & 0.375 & \makecell{0.484\\{[}0.407, 0.586{]}} & \makecell{0.742\\{[}0.645, 0.863{]}} & \makecell{0.798\\{[}0.740, 0.864{]}} & \makecell{0.921\\{[}0.875, 0.958{]}} & \makecell{0.725\\{[}0.590, 0.855{]}} & 21/36 & 0.638 \\[2pt]
Random tiles (no scout) & none & no & 0.375 & \makecell{0.599\\{[}0.544, 0.657{]}} & \makecell{0.685\\{[}0.603, 0.807{]}} & \makecell{0.767\\{[}0.709, 0.847{]}} & \makecell{0.921\\{[}0.884, 0.952{]}} & \makecell{0.697\\{[}0.602, 0.791{]}} & 0.563 & 0.706 \\[2pt]
\bottomrule
\end{tabular}
\par\medskip
\begin{minipage}{\textwidth}\footnotesize Native decision of SASHA's own classifier (HAFED), for comparison: on all cells, pooled AUROC 0.931 [0.892, 0.961] and ITC AUROC 0.753 [0.651, 0.852], with 22/36 ITC slides called; at b20, 0.867 [0.819, 0.909] and 0.563 [0.453, 0.676], with 11/36 called.\end{minipage}
\end{sidewaystable}

The three seeds agreed on the direction of the three primary contrasts (Fig. 2b), but whether SASHA's advantage over random cells was established depended on the seed. SASHA's ITC capture at b20 was 0.495, 0.463 and 0.589 in seeds 4, 5 and 6, against 0.342, 0.296 and 0.326 for random cells and 0.884, 0.853 and 0.874 for attention top-R. Its advantage over random cells was established in seeds 5 and 6 (0.167 [0.055, 0.288] and 0.263 [0.152, 0.383]) but not in seed 4, whereas its gap to attention was established in all three seeds ($-$0.389, $-$0.389 and $-$0.284). Random tiles had higher point estimates than SASHA in seeds 4 and 5 and about the same in seed 6 (0.599, 0.558 and 0.583). None of these differences was established; in seed 4 the difference was $-$0.104 [$-$0.258, 0.049]. In seed 4, SASHA's own checkpoint rule selected policies from epoch 0 at b20 in two folds, and the held-out centres of those folds lost lesion capture (all categories) from b10 to b20, from 0.439 to 0.241 and from 0.588 to 0.502. Replaying the policies from the last epoch post hoc raised ITC capture to 0.516 [0.384, 0.656], with an established advantage over random cells (0.188 [0.063, 0.312]) and a gap to attention of $-$0.347 [$-$0.500, $-$0.174].

\subsection{Slide decisions}

\textbf{Native decisions.} SASHA's own classifier lost little pooled discrimination but much ITC discrimination (Table 2, note). Against HAFED on all cells, SASHA at b20 had a pooled AUROC of 0.867 against 0.931 (difference $-$0.063 [$-$0.099, $-$0.031]) but an ITC AUROC of 0.563 against 0.753 ($-$0.190 [$-$0.324, $-$0.054]). The ITC loss thus exceeded the pooled loss by 0.126 [0.019, 0.237] (R-D: established; within centre, 0.157 [0.057, 0.258]). SASHA called 11 of 36 ITC slides at the validation threshold, against 22 for HAFED. Across the three seeds, the deficit in ITC AUROC was $-$0.190, $-$0.105 [$-$0.208, $-$0.018] and $-$0.059 [$-$0.165, 0.040], while pooled AUROC stayed within 0.02--0.06 of the full bag. The direction held in every seed, but in seed 6 the interval included 0, as did the interval of the excess over the pooled loss (0.039 [$-$0.039, 0.124]; seed 5, 0.076 [0.006, 0.161]).

\textbf{Common predictor.} Under the frozen full-bag ABMIL at the b20 cost of SASHA seed 4, the full bag reached an ITC AUROC of 0.779 (Table 2; Fig. 2c). SASHA's tiles reached 0.608 (difference $-$0.171 [$-$0.302, $-$0.052]), scout + random cells 0.615 ($-$0.163 [$-$0.215, $-$0.109]) and random tiles 0.697 ($-$0.081 [$-$0.120, $-$0.045]), while scout-greedy reached 0.725 ($-$0.054 [$-$0.147, 0.026]). The pooled AUROC of the achievable policies fell by only 0.021--0.063, and their macrometastasis AUROC stayed between 0.993 and 1.000. The decision loss was therefore not general across achievable policies (D1: no). Scout-greedy, which is supervised with lesion labels, was the exception in seed 5 as in seed 4, and in seed 6 SASHA's tiles were a second exception ($-$0.054 [$-$0.119, 0.010]). The ITC loss exceeded the pooled loss by 0.108 [0.015, 0.218] for SASHA's tiles, 0.105 [0.061, 0.149] for random cells and 0.060 [0.032, 0.091] for random tiles (R-D: established), but the excess was not established for scout-greedy (0.033 [$-$0.032, 0.108]). Counting only pairs within a held-out centre gave 0.087, 0.093, 0.056 and 0.028, with the same conclusions. Outside the prespecified reading, every achievable policy lost ITC discrimination at b10, scout-greedy included ($-$0.103 [$-$0.206, $-$0.012]). SASHA's loss was not only a property of its own classifier. It persisted under the common predictor in seeds 4 and 5 (D2: yes; seed 5, $-$0.094 [$-$0.177, $-$0.021]) and was not established in seed 6. No difference in ITC AUROC between SASHA's tiles and any comparator was established (D3), so we state no ranking. At T1, random tiles (+0.202 [0.072, 0.347]) and scout-greedy (+0.222 [0.030, 0.432], at a specificity cost of $-$0.082 [$-$0.117, $-$0.045]) called more ITC slides than SASHA's tiles in seed 4, but neither difference was established in seeds 5 and 6. Transferring the full-bag threshold (T2) moved the operating point without recovering ITC calls: SASHA's tiles then called 12 of 36 ITC slides at a specificity of 0.846.

\textbf{Evidence versus call.} On the 29 assessable ITC slides (seed 4, b20, T1), the full bag had the evidence of every slide available and called 21 of them. Paired slide by slide under the common predictor, SASHA's tiles kept 11 of these calls, lost 10 and gained none. Seven of the lost calls were on slides where SASHA acquired no annotated lesion, and 3 were on slides where it captured one. SASHA's native classifier also missed all 10 slides without an acquired lesion; both full-bag classifiers called 7 of them, and neither called the other 3. Under the common predictor at b20, most lost calls fell on slides without an acquired lesion in every seed (7 of 10, 4 of 6 and 2 of 3), but at b10 half did (5 of 10). Against HAFED, SASHA's own classifier showed the same pattern in seeds 4 and 5 (7 of 10 and 5 of 9 lost calls without an acquired lesion) but not in seed 6, where 4 of its 6 lost calls fell on slides with a captured lesion and it gained 3 calls, 2 of them on slides without one (Supplementary Table S4b).

\subsection{Robustness of the lesion endpoint}

The capture gap to attention survived both alternative lesion rules (L1: yes). For lesions on ITC slides at b20, SASHA minus attention top-R was $-$0.379 [$-$0.588, $-$0.167] under substantial overlap and $-$0.338 [$-$0.480, $-$0.204] with merged units, which turn the 95 lesions on ITC slides into 71 units, and the gap stayed below 0 at every budget and seed. Random tiles depended most on the overlap rule. Their ITC capture fell from 0.599 to 0.427 under substantial overlap, whereas SASHA's fell from 0.495 to 0.463. Their lead over SASHA in point estimates thus held under any overlap and with merged units (0.636 against 0.549) but reversed under substantial overlap, although random tiles exceeded random cells under every rule. Selector loss, not grid loss, dominated: counting only lesions that overlap at least one candidate tile, ITC capture at b20 was 0.511 for SASHA and 0.913 for attention top-R.

Excluding the doubtful slide patient\_068\_node\_1 left every prespecified reading unchanged in all four audits (seeds 4, 5 and 6 and the last-epoch run; reading E1). In seed 4, SASHA minus attention became $-$0.411 [$-$0.592, $-$0.213]. We also deleted one slide or one case at a time. The contrasts against random cells and against attention kept their sign under every deletion in all four audits. Only the contrast of SASHA minus scout-greedy, whose point estimate was close to 0 (0.011--0.032), changed sign. The prespecified influence reading (E2, no reversal of any primary contrast) was therefore ``no'' for seeds 4 and 5 and the last-epoch run, and ``yes'' for seed 6. In the whole-cell variant of the random-cell comparator, the last cell overshot $R$ by 0.0013 of $N$ on average and changed capture and AUROC by less than 0.01, so cutting the last cell did not drive the results for random cells. Summaries at slide level swapped the order of SASHA and scout-greedy: their ITC capture balanced over slides was 0.550 and 0.574, and the share of ITC slides with any lesion captured was 0.655 (19 of 29) and 0.793. These summaries did not change the gap to attention, which reached 0.888 and 0.931.

\subsection{Calibrated fallback on development folds}

The tested cascade driven by SASHA's score did not meet its release target in two held-out centres (Fig. 3; full cascade tables in Supplementary Table S7). At a mean cost of 0.654 [0.620, 0.686], it released 16 of 98 small-lesion slides as negative (0.163 [0.096, 0.252]). Fifteen of the released slides were ITC slides, and the full pipeline would have called 9 of the 16. Point estimates were within $\alpha$ in 3 of 5 centres; centre 0 released 8 of 21 small-lesion slides, and centre 4 released 6 of 13. The prespecified controlled saving was therefore not achieved. On several released slides, SASHA had acquired none of the annotated lesions. An oracle fallback that knew which positives the cheap pass would miss would have cost 0.434 with no added miss.

The cascade driven by the scout score released 2 of 98 small-lesion slides (0.020 [0.002, 0.072]), with point estimates within $\alpha$ in all five centres, at a cost of 0.557 [0.515, 0.598]. It called 25 of 36 ITC slides (full pipeline, 23), with 2 added misses against the full pipeline. Its specificity, however, fell from 0.720 to 0.531 (169 of 318 negatives), and to 13 of 75 and 10 of 61 in centres 2 and 4, where most of the saving occurred. Part of the saving thus moved work from encoding to the review of false positives. The release-only variants separated the two mechanisms. For the scout cascades that met the release target, release alone cost 1.043 [1.022, 1.062] of a full encode (R1: no). Only the SASHA cascades, which released 14--16 of the 98 small-lesion slides, saved encoder passes through release alone (0.907--0.913).

\subsection{One-time confirmation}

On the 858 held-out slides, the frozen gatekeeper met all three primary criteria (Table 3). For P1, 1 of 130 small-lesion slides was released (0.008 [0.000, 0.042]). For P2, the mean cost fraction was 0.836 [0.800, 0.871]. For P3, the specificity difference was $-$0.014 [$-$0.026, $-$0.003] (0.821 against 0.835). No ITC slide was released (0 of 18), but with an upper bound of 0.185 the result is not specific to ITC, because P1 pools 18 ITC and 112 micrometastatic slides. The single release was a CAMELYON16 micrometastatic slide that the full pipeline had called, and it was the only added miss. Ten negatives became new false positives through positive stops, and two false positives of the full pipeline were rescued by release, so false positives rose from 164.6 to 179.0 per 1,000 negatives. At case level, 1 of 113 cases with a small-lesion slide had a slide released [0.000, 0.048], so the case view did not contradict P1 (reading C1).

The saving came from early positive stops. Of 175 macrometastatic slides, 153 stopped at the scout, as did 31 of 112 micrometastatic slides, and macrometastatic slides cost 0.218 of a full encode on average. Negatives cost 1.031, more than a full encode, because 517 of 553 negatives fell back and also paid for the scout. Only 15 slides were released, 14 of them from CAMELYON16, and all 15 fell in the smallest quartile of cell count. Over the cohort, the ratio of total cost to total $N$ was 0.790. The secondary one-sided variant, which stops at the validation Youden threshold, also met all three primary criteria, at a higher cost and with a smaller loss of specificity (cost 0.859 [0.823, 0.894]; specificity difference $-$0.002 [$-$0.010, 0.006]). On this set, the stop threshold calibrated with $\beta$ did not protect specificity better than the Youden threshold. No confirmation score tied with any threshold.

\begin{table}[!t]
\centering
\small
\caption{The frozen scout gatekeeper, run once on the 858 held-out slides from the same hospital family. P1--P3 were fixed before any confirmation score was computed; the other rows are secondary or post hoc and do not change the verdict.}
\label{tab:3}
\begin{tabularx}{\textwidth}{@{}>{\raggedright\arraybackslash}X>{\raggedright\arraybackslash}p{2.7cm}>{\raggedright\arraybackslash}p{4.7cm}>{\raggedright\arraybackslash}p{1.7cm}@{}}
\toprule
Quantity & Criterion & Estimate [95\% CI] & Met \\
\midrule
P1: small-lesion (ITC + micro) slides released & upper bound $\leq$ 0.10 & 1/130 = 0.008 [0.000, 0.042] & yes \\
P2: mean per-slide cost / $N$ & upper bound < 0.90 & 0.836 [0.800, 0.871] & yes \\
P3: specificity, cascade $-$ full pipeline & lower bound $\geq$ $-$0.05 & $-$0.014 [$-$0.026, $-$0.003] (0.821 vs 0.835) & yes \\
ITC slides released & --- & 0/18 [0.000, 0.185] & not ITC-specific \\
Cases with a released small-lesion slide & --- & 1/113 [0.000, 0.048] & --- \\
Added misses; rescued positives & --- & 1 (micro, CAMELYON16); 0 & --- \\
False positives per 1,000 negatives, full pipeline $\rightarrow$ cascade & --- & 164.6 $\rightarrow$ 179.0 & --- \\
Total cost / total $N$ & --- & 0.790 & --- \\
Mean cost / $N$: negative; ITC; micro; macro & --- & 1.031; 0.988; 0.810; 0.218 & --- \\
Released / stopped / fell back: negative; ITC; micro; macro & --- & 14/22/517; 0/2/16; 1/31/80; 0/153/22 & --- \\
One-sided variant: P1; P2; P3 & as above & 1/130; 0.859 [0.823, 0.894]; $-$0.002 [$-$0.010, 0.006] & yes \\
\bottomrule
\end{tabularx}
\end{table}

\begin{figure}[!t]
\centering
\includegraphics[width=\textwidth]{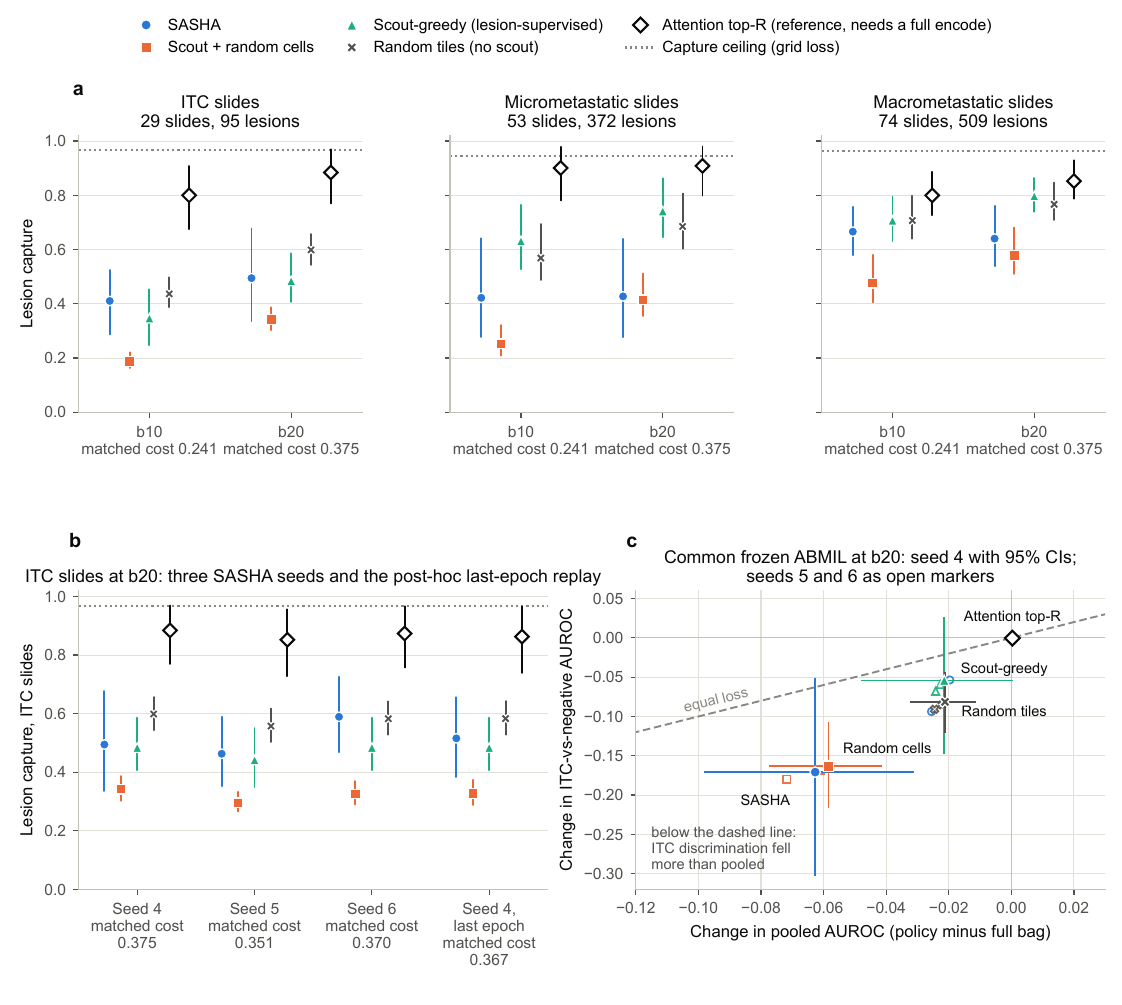}
\caption{Acquisition trade-off at matched total cost. (a) Lesion capture on ITC, micrometastatic and macrometastatic slides at the b10 and b20 budgets (SASHA seed 4), with 95\% CIs. The matched cost under each budget is SASHA's mean cost, 0.241 and 0.375 of a full encode, to which every achievable policy is matched; attention top-R needs a full encode (cost 1.000) and is a reference, not a cost-saving policy. The dotted line is the capture ceiling set by grid loss. (b) Capture on ITC slides at b20 for the three SASHA seeds and the post hoc replay of the last-epoch policies; comparator values differ between settings because each comparator is matched to the cost of its setting. (c) Change in pooled AUROC against change in ITC AUROC (ITC against negative slides), relative to the full bag, under the common frozen ABMIL at b20: seed 4 with 95\% CIs, seeds 5 and 6 as open markers. Points below the dashed identity line lost more ITC discrimination than pooled discrimination. The bars are marginal intervals; the paired interval for the distance below the line, D, is given in Section 4.4 (for SASHA's tiles, 0.108 [0.015, 0.218]).}
\label{fig:2}
\end{figure}

\begin{figure}[!t]
\centering
\includegraphics[width=\textwidth]{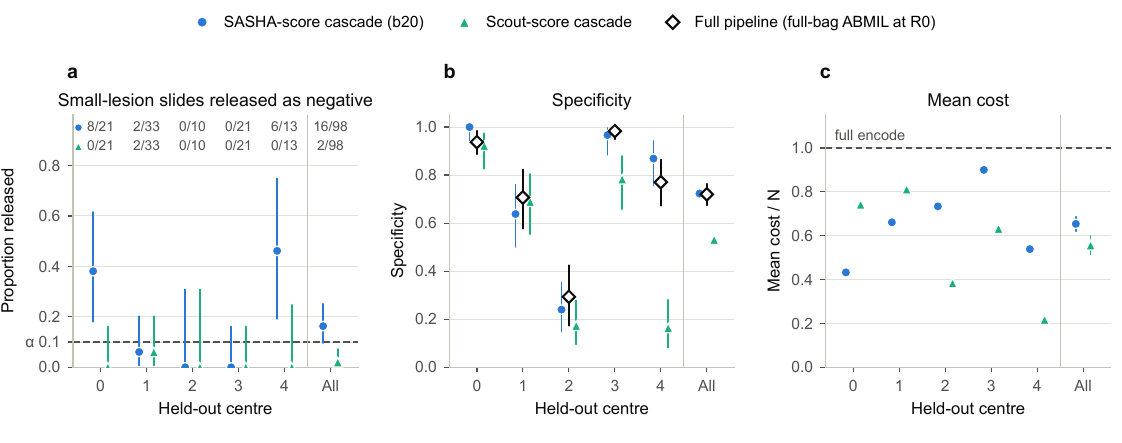}
\caption{The release and fallback cascades on development folds, by held-out centre. (a) Proportion of small-lesion (ITC and micrometastatic) slides released as negative, with exact 95\% Clopper--Pearson intervals; the counts give released over small-lesion slides for the cascades driven by SASHA's score (upper row) and by the scout score (lower row), and the dashed line marks $\alpha$ = 0.10. (b) Specificity of each cascade (exact intervals) and of the full pipeline, full-bag ABMIL at R0 (bootstrap intervals over cases). (c) Mean cost per slide as a fraction of a full encode; the dashed line marks a full encode. ``All'' pools the five held-out centres, with a bootstrap interval over cases for its cost.}
\label{fig:3}
\end{figure}

\section{Discussion}

\subsection{Principal finding}

Cost-saving acquisition lost ITC evidence and ITC discrimination far more than pooled slide metrics showed. SASHA, a recent learned policy, lost about 0.06 of pooled AUROC against its classifier on the full slide, while its ITC AUROC fell by 0.19 and it captured 0.495 of the lesions on ITC slides, against 0.884 for the attention reference. The excess of the ITC loss over the pooled loss, 0.126 [0.019, 0.237], had a paired interval above 0. The disparity is not specific to SASHA. Under one frozen predictor at the same cost, random cells and random tiles showed it too, with excesses of 0.105 and 0.060 whose intervals lay above 0. Nor is it universal. The scout policy supervised with lesion labels did not establish an ITC loss at b20, and in one of three SASHA seeds the excess was not established. Pooled AUROC understates the ITC disparity across the slide-supervised and unsupervised policies we tested, and an evaluation that reports only pooled metrics cannot tell which kind of policy it is looking at.

\subsection{Relation to prior work}

BagShift showed, with stressor selectors applied to cached bags, that equal patch counts can preserve unequal evidence and that missed micrometastatic tissue drove the largest response on CAMELYON16~\cite{yuan2026bagshift}. Our results extend that observation to learned acquisition at matched total encoding cost, to lesions as units on ITC slides and to held-out centres, and the two studies agree in direction. Dispersed random tiles kept more lesions on ITC slides than contiguous random cells under every lesion rule, just as BagShift's dispersed budget kept more tumour than its spatial window; we did not test this contrast, and the lead of random tiles over SASHA depended on the overlap rule. BagShift also showed that micrometastasis discrimination falls more than macrometastasis discrimination under selection (0.985 to 0.791 against 1.000 to 0.994 at its random budget of 128 patches), and that strong ranking of micrometastases can coexist with a poor operating point. A subgroup loss larger than the pooled loss is therefore not new. What our data add is its size for ITC when acquisition is learned or guided by the scout, at matched total cost and with centres held out.

Against a common set of negatives, pooled AUROC is the mean of the subtype AUROCs weighted by their numbers of positives. ITC slides carry 36 of the 174 positive slides here, so a large ITC loss moves pooled AUROC little, which is the hidden stratification problem of aggregate metrics~\cite{oakdenrayner2020hidden}. Sensitivity at lesion level by metastasis category and held-out centre has also been reported for a segmentation model on CAMELYON17~\cite{almobayed2026pyramid}. Our endpoint differs in asking what a budgeted acquisition retains before any classifier sees it.

SASHA reported parity at slide level with methods that use every patch, including on four-class CAMELYON+ classification~\cite{gogisetty2025sasha}. Its authors noted that at a 10\% budget SASHA tends to misclassify slides with small tumour regions, and that at 20\% it recovered nearly all slides that HAFED classified correctly on binary CAMELYON16 labels. On ITC slides at b20, SASHA called 11 of 36 against 22 for HAFED. Our view by subtype does not contradict their pooled report, since pooled AUROC fell only modestly here as well, but it shows what a pooled report leaves out. HIPPO found limitations of metastasis detection models that standard metrics and attention did not reveal~\cite{kaczmarzyk2024hippo}, and lesion capture at matched cost is a complementary probe on the acquisition side.

\subsection{Evidence loss and decision loss}

Two failures can lower the ITC decision of a cost-saving policy: not acquiring the lesion, and not calling a slide whose lesion was acquired. The paired transition tables show how often each failure accompanied a lost call, although thresholded calls cannot decompose a loss of AUROC. Under the common predictor, SASHA's tiles lost 10 of the full bag's 21 ITC calls and gained none. Of the lost calls, 7 were on slides where SASHA acquired no annotated lesion and 3 on slides where it captured one. At b20, most lost calls under this predictor fell on slides without an acquired lesion in every seed, so missing evidence accompanied most of them, and the ITC loss itself persisted in two of three seeds. Under SASHA's own classifier the same held in seeds 4 and 5 but not in seed 6, where 4 of 6 lost calls fell on slides with a captured lesion. At b10, half of the lost calls under the common predictor also fell on such slides. Acquiring a lesion was not always enough, and how much of the ITC deficit goes with missing evidence depends on the reader and the seed.

Full information shows that the evidence is reachable: attention top-5\% kept annotated tumour on 27 of 29 assessable ITC slides and kept pooled discrimination. Selection can also change a call in the other direction, since attention top-5\% recovered two ITC slides that the full bag missed. Six ITC slides, however, scored below 0.07 on the full bag and were never called by the full bag or by attention selection under any threshold rule. Random selections called them only in a minority of draws at the validation threshold, or under thresholds that called almost every slide (specificity 0.015 or 0.072). The checkpoint rule also matters. Two policies from epoch 0 lowered capture at b20, and policies from the last epoch raised ITC capture on average, although not in every fold. In seeds 4 and 5, scout-greedy, which is supervised with lesion labels, was the only cheap policy without an established ITC loss at b20 (in seed 6, SASHA's tiles were a second), which suggests that supervising the acquisition signal at lesion level may help to protect ITC evidence. This is a hypothesis: the comparison is not like-for-like in supervision, and scout-greedy itself lost ITC discrimination at b10.

\subsection{What evaluations of acquisition should report}

The audit suggests a minimum reporting set for cost-saving whole-slide analysis:

\begin{itemize}
\item lesion capture stratified by metastasis category, at matched total cost that includes the scout and the selection signal;
\item grid loss reported separately from selector loss;
\item pooled AUROC and AUROC by subtype, under the policy's own classifier and under a common predictor;
\item release and miss counts per held-out centre, with denominators;
\item prespecified readings, and the supervision level of every comparator;
\item the checkpoint rule and the number of training seeds.
\end{itemize}

For cohorts like CAMELYON, which already provide lesion annotations and centre labels, none of these items requires new data.

\subsection{The gatekeeper trade-off}

A calibrated release rule transfers only as far as its cheap score keeps small-lesion slides above the development threshold in each new centre. With SASHA's score, the tested cascade did not meet its target in two of five centres. The scout gatekeeper met its three frozen criteria once on held-out slides from the same hospital family, but its saving was modest (0.836 per slide on average, 0.790 of the cohort's encoding) and came from early stops on macrometastatic slides. Negative release was confined to the smallest slides, because a maximum over cells tends to grow with the number of cells (Spearman correlation 0.610 between the scout maximum and the cell count in negatives). On development folds, negative release saved no encoder passes on its own. On the confirmation set, the stop threshold calibrated with $\beta$ did not protect specificity better than the Youden threshold of the one-sided variant. Requiring all five pipelines to agree made positive stops on negatives rare (22 of 553), and the full pipeline was more specific here (0.835) than on the benchmark's held-out centres (0.720). Both are plausible reasons why P3 passed despite our expectation from development, and neither was tested. The confirmation is neither specific to ITC nor a test on a new hospital.

\subsection{Limitations}

The cases are constructed challenge cases, not verified patients, and every slide comes from one hospital family; five centres are too few for inference about variation between hospitals. The release provides no demographic data, so sex, age and other patient characteristics could not be examined. We audited only one learned method, through an adapter to the release's tile grid and features rather than in SASHA's original pipeline, and on binary labels without the ITC oversampling of its four-class variant. A policy trained with that oversampling might acquire more ITC evidence. Only 29 ITC slides could be assessed at lesion level, and the intervals are wide. The endpoint is geometric and relies on annotations that are not exhaustive, were not adjudicated again and were read under the publisher's group semantics. The correspondence between feature rows and coordinates was verified on samples only. Costs count encoder passes, not wall-clock time, input and output, or policy computation. Scout-greedy uses lesion labels, so it is a comparator with stronger supervision. The common predictor was trained on full bags, and a predictor retrained on subsets might lose less. No pathologist reviewed the slides that most influenced the results. Finally, the benchmark test folds were read repeatedly, so every benchmark result is a discovery result; only the single confirmation of the gatekeeper is confirmatory.

\subsection{Future work}

The next steps are a second family of learned acquisition methods, an external lymph node cohort with lesion annotations, a timing panel on raw slides that measures wall-clock cost together with encoder passes, and a blinded pathology review of influential slides such as patient\_068\_node\_1 and patient\_081\_node\_4. Acquisition objectives that use lesion labels during development, which our scout-greedy result points to, should be compared with deployable policies, random sampling included, at matched total encoder passes, and with the attention reference at a matched count of retained tiles.

\section{Conclusion}

On a CAMELYON17 benchmark of five centres with annotated reference lesions, cost-saving acquisition at 0.375 of the full encoding cost lost lesion evidence on ITC slides, and ITC discrimination, far more than its pooled slide AUROC suggested. A learned policy, SASHA, captured 0.495 of the lesions on ITC slides, against 0.884 for the attention reference. Under one frozen predictor at the same cost, the ITC loss exceeded the pooled loss for SASHA's tiles and both random comparators in the primary seed, with post hoc paired intervals above 0; for a scout policy supervised with lesion labels, the ITC loss was not established. A calibrated fallback driven by SASHA's score did not meet its release target in two of five held-out centres. A scout gatekeeper trained with lesion labels met three frozen criteria once on held-out slides from the same hospital family, saving encoder passes mainly through early stops on macrometastases. These are results on constructed challenge cases and annotated reference lesions, not clinical results. Evaluations of cost-saving whole-slide acquisition should report lesion capture by metastasis category at matched total cost, alongside pooled slide metrics.

\section*{CRediT authorship contribution statement}

\textbf{Iftikhar Ahmed:} Conceptualization, Methodology, Software, Validation, Formal analysis, Investigation, Data curation, Visualization, Writing -- original draft, Writing -- review and editing, Project administration.

\textbf{Adiba Rahman Namira:} Investigation, Validation, Data curation, Writing -- review and editing.

\section*{Declaration of competing interest}

The authors declare that they have no known competing financial interests or personal relationships that could have appeared to influence the work reported in this paper.

\section*{Funding}

This research did not receive any specific grant from funding agencies in the public, commercial, or not-for-profit sectors.

\section*{Ethics statement}

This study used only publicly available, de-identified whole-slide images and annotations from the CAMELYON16, CAMELYON17 and CAMELYON+ releases~\cite{litjens2018camelyon,ling2025camelyonplusdata}. No new human data or tissue were collected, and no person was contacted. Formal ethical approval was therefore not sought.

\section*{Data and code availability}

The slides are public on the Registry of Open Data on AWS (CC0)~\cite{camelyonaws}, and the corrected labels, annotations, coordinates and UNI features are available from the CAMELYON+ release~\cite{ling2025camelyonplusdata}. The UNI weights are gated by their owners and cannot be redistributed~\cite{uni_model}. The benchmark assignment table (492 rows; SHA-256 17a991b04f12d6ef810d44a73019264569bd4694c095f7ab1d806b6d9ff00f67) and the SHA-256 hashes of every derived table are listed in the analysis records. The analysis code, the SASHA adapter patch and the scripts that build the figures and tables are available from the corresponding author on reasonable request.

\section*{Declaration of generative AI and AI-assisted technologies in the manuscript preparation process}

All references were obtained via Google Scholar search, and we studied them ourselves. We used Claude Code (Opus 5.5) for minor formatting and code scaffolding during paper preparation then we manually reviewed it. After using this tool, the authors reviewed and edited the content as needed and take full responsibility for the content of the published article.

\section*{Acknowledgements}

The authors thank the organisers of the CAMELYON16 and CAMELYON17 challenges, the creators of the CAMELYON+ release, and the developers of UNI and SASHA for making their data, model and code publicly available.

\clearpage
\includepdf[pages=-]{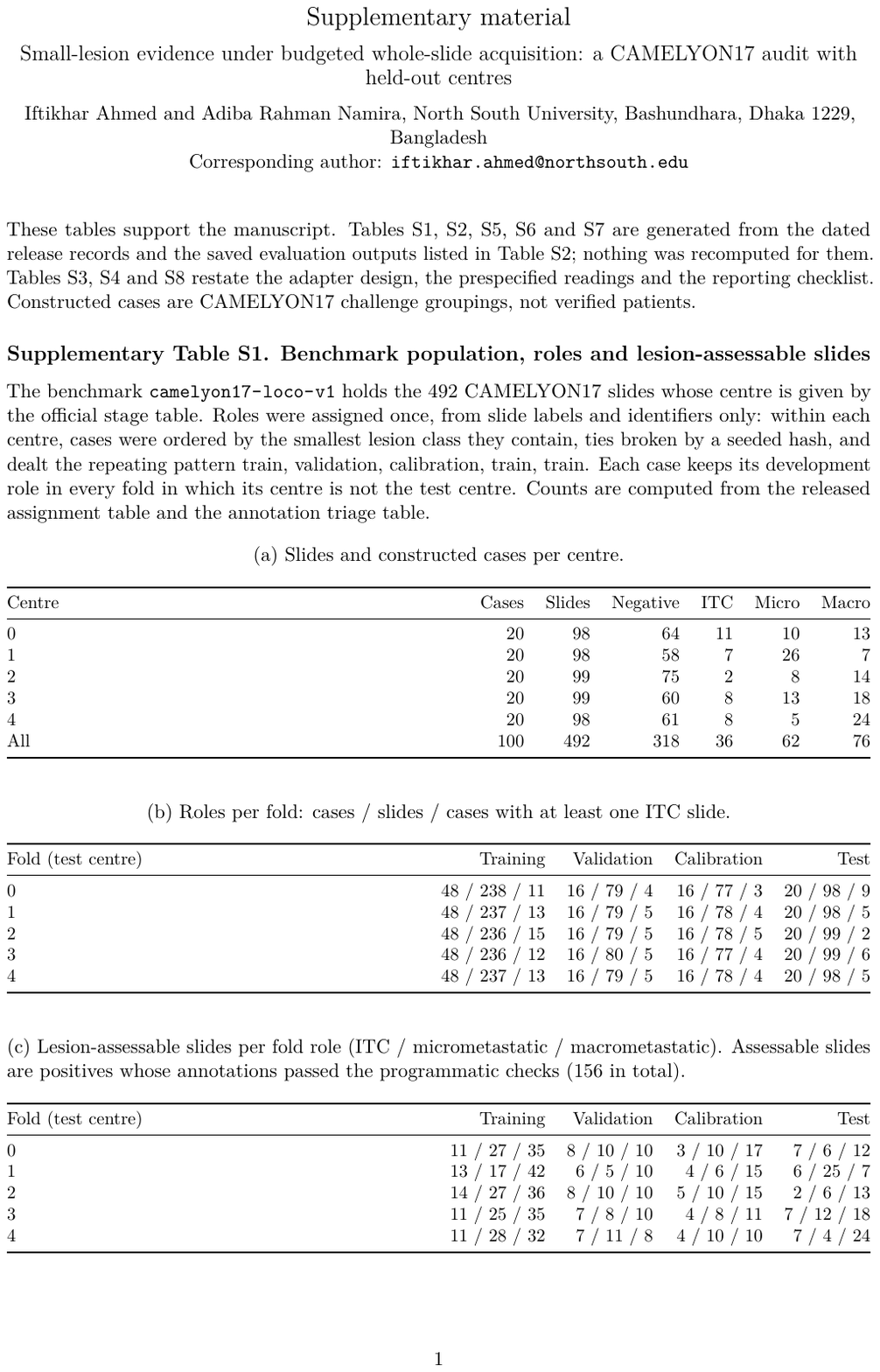}

\end{document}